\documentclass[10pt,twocolumn,letterpaper]{article}

\usepackage[pagenumbers]{cvpr} 

\usepackage{graphicx}
\usepackage{amsmath}
\usepackage{amssymb}
\usepackage{booktabs}
\usepackage{multirow}

\usepackage{enumitem}
\usepackage[T1]{fontenc}
\usepackage{mathabx,graphicx}

\usepackage[dvipsnames]{xcolor}

\newcommand{\high}[1]{{\textbf{\color[RGB]{0,0,255}#1}}}
\newcommand\Tstrut{\rule{0pt}{2.3ex}}

\usepackage[pagebackref,breaklinks,colorlinks]{hyperref}

\usepackage[capitalize]{cleveref}
\crefname{section}{Sec.}{Secs.}
\Crefname{section}{Section}{Sections}
\Crefname{table}{Table}{Tables}
\crefname{table}{Tab.}{Tabs.}

\def\cvprPaperID{7410} 
\def\confName{CVPR}
\def\confYear{2022}

\begin{document}

\title{HOP: History-and-Order Aware Pre-training for\\
Vision-and-Language Navigation}

\author{
Yanyuan Qiao$^{1}$ \quad Yuankai Qi$^1$ \quad Yicong Hong$^2$ \quad Zheng Yu$^{1}$ \quad Peng Wang$^3$ \quad Qi Wu$^{1}$\thanks{Corresponding author}\\
$^1$The University of Adelaide ~~ $^2$The Australian National University ~~\\ $^3$Northwestern Polytechnical University\\
{\tt\small \{yanyuan.qiao,qi.wu01\}@adelaide.edu.au, \{qykshr,william.zhengyu\}@gmail.com} \\ 
{\tt\small yicong.hong@anu.edu.au, peng.wang@nwpu.edu.cn } \\
{\tt\small \href{https://github.com/YanyuanQiao/HOP-VLN}{https://github.com/YanyuanQiao/HOP-VLN}}
}


\maketitle

\vspace{-20pt}
\begin{abstract}
Pre-training has been adopted in a few of recent works for Vision-and-Language Navigation (VLN). However, previous pre-training methods for VLN either lack the ability to predict future actions or ignore the trajectory contexts, which are essential for a greedy navigation process. In this work, to promote the learning of spatio-temporal visual-textual correspondence as well as the agent's capability of decision making, we propose a novel history-and-order aware pre-training paradigm (HOP) with VLN-specific objectives that exploit the past observations and support future action prediction. Specifically, in addition to the commonly used Masked Language Modeling (MLM) and Trajectory-Instruction Matching (TIM), we design two proxy tasks to model temporal order information: Trajectory Order Modeling (TOM) and Group Order Modeling (GOM). Moreover, our navigation action prediction is also enhanced by introducing the task of Action Prediction with History (APH), which takes into account the history visual perceptions. Extensive experimental results on four downstream VLN tasks (R2R, REVERIE, NDH, RxR) demonstrate the effectiveness of our proposed method compared against several state-of-the-art agents.

\end{abstract}
\vspace{-20pt}

\section{Introduction}
\label{sec:intro}
\vspace{-5pt}
Vision-and-Language Navigation (VLN) has received large attention in communities of computer vision, natural language processing and robotics due to its great importance towards real-world applications such as domestic assistants~\cite{r2r,rcm, speakerfollower,touchdown,hanna,vlnce,VisionBase_imi}.
VLN requires an agent to navigate to a target location in a 3D simulated environment, according to a given natural language instruction. 
In the past few years, a great variety of VLN tasks have been proposed, including navigation with low-level instructions such as R2R~\cite{r2r} and RxR~\cite{rxr}, communicative and cooperative instructions such as NDH~\cite{ndh}, and high-level instructions for remote object grounding such as REVERIE~\cite{reverie} and SOON~\cite{soon}.

Despite their differences, the agent's navigation is mostly formulated as a sequential text-to-image grounding problem.
That is, positioned at a particular node on a pre-defined connectivity graph, the agent traverses the environment by selecting the adjacent node that has the maximum correspondence between the image representation and the instruction. 
As a result, the visual-textual matching is considered to be the keystone of addressing VLN tasks.

\begin{figure}[!t]
	\begin{center}
		\includegraphics[width=1.0\linewidth]{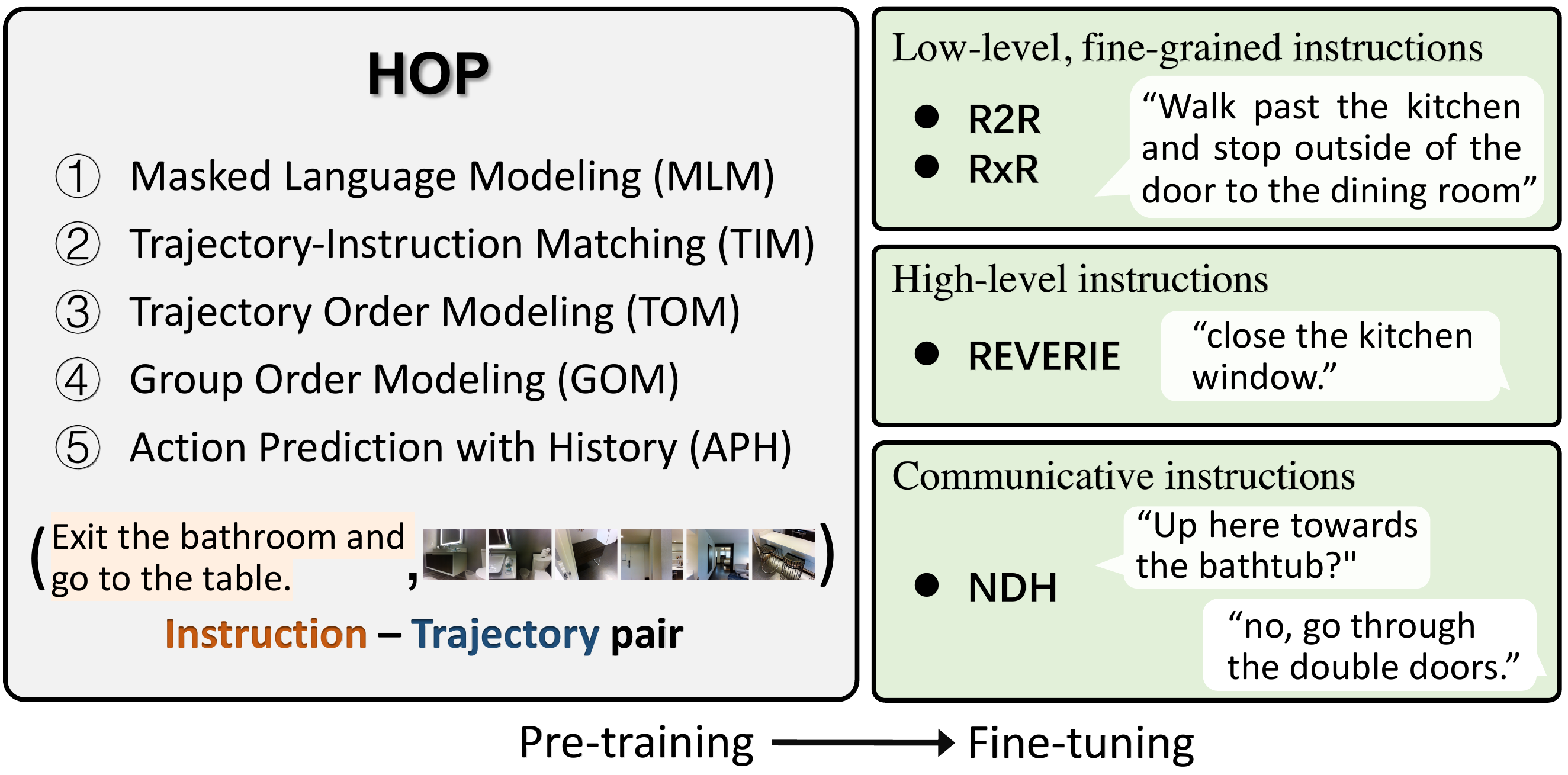}
	\end{center}
	\vspace{-17pt}
	\caption{{Illustration of the proposed pre-training and ﬁne-tuning paradigm for VLN. The model is pre-trained with five proxy tasks, and fine-tuned on four downstream VLN tasks: R2R, RxR, REVERIE and NDH (detailed in Section~\ref{sec:HOP model})}.
	}
	\label{fig:pretrain}
	\vspace{-17pt}
\end{figure}

Inspired by the great success of Vision-Language BERT pre-training on several visual-textual matching tasks, such as image-text retrieval~\cite{itr} and referring expression grounding~\cite{reg}, several pre-training methods have been proposed for VLN~\cite{prevalent,altr,bertvln,airbert}.
These approaches are able to achieve better performance, but they still suffer from some limitations.
VLN-BERT~\cite{bertvln} pre-trains its model by predicting the compatibility of a pair of instruction and visual trajectory. In the downstream tasks, it formulates the navigation as a trajectory selection problem. AirBERT~\cite{airbert} further adopts a binary classification task to predict whether the given instruction and visual trajectory are paired. 
Both VLN-BERT and AirBERT discard navigating action prediction during pre-training, weakening the relation between the learned representation and the final goal: navigation action prediction. 
By contrast, PREVALENT~\cite{prevalent} introduces a single-step action prediction task, aiming to learn action-oriented generic visiolinguistic representation, which can be applied to the greedy search VLN. 
However, PREVALENT largely overlooked the important historical context in pre-training. {It only takes the static panoramic image of a single step as visual input, while failing to take into account the history trajectory information.}
Indeed, VLN is a Partially Observable Markov Decision Process (POMDP), where the agents rely heavily on the past experiences for making future action decisions. {Furthermore, VLN is a spatio-temporal task which is sensitive to the sequence order of the trajectory. Thus the ability of temporal order reasoning is also beneficial to the action decision making. Nevertheless, all the above three methods do not explicitly mine temporal order information from either instructions or visual observations.}

To address the above mentioned issues, in this work, we propose a novel {history-and-order aware} pre-training paradigm to enhance the learning of visual-textual correspondence for VLN task.
\textbf{First}, we provide history visual observations to the action prediction task, called Action Prediction with History (APH), which helps the model locate the sub-instruction to be executed and thus improve the action prediction accuracy.
\textbf{Second}, we design two order-aware proxy tasks, Trajectory Order Modeling (TOM) and Group Order Modeling (GOM). Given an instruction, TOM requires the model to recover the order of shuffled visual trajectory from a fine-grained level, and GOM requires the model to predict the order of two groups of sub-trajectories from a coarse level.
These two tasks explicitly equip the model with the ability to understand the temporal order within instructions, in addition to the visual-textual matching capability. 
The overall of the proposed pre-training and ﬁne-tuning tasks are illustrated in Figure~\ref{fig:pretrain}.

To comprehensively evaluate our proposed pre-training methods, we conduct experiments on four downstream tasks:  R2R~\cite{r2r}, RxR~\cite{rxr}, NDH~\cite{ndh}, REVERIE~\cite{reverie}. Each task poses a very different challenge to evaluate the agent. R2R serves as an in-domain task, which can verify the agent's generalization ability to unseen environments. The other three tasks are out-of-domain, which are used to study the generalization ability to new tasks. RxR is known for longer instructions. NDH features dialog instructions. REVERIE is characterized by high-level, short instructions. With our proposed pre-training tasks, the fine-tuned downstream model performs favorably on all these tasks: 59\% SPL on R2R,  0.33 sDTW on RxR, 3.31 GP on NDH, and 24.34\% SPL (14.34\% RGSPL) on REVERIE.
\section{Related work}
\label{sec:related_work}
In this section, we briefly review several closely related works in VLN and Vision-Language pre-training.

\vspace{-7pt}
\paragraph{Vision-and-Language Navigation}
Vision and language navigation task has attracted a lot of attention since it was proposed in the room-to-room task~\cite{r2r}. 
This task is initially cast as a vision-based sequence-to-sequence trans-coding problem~\cite{r2r}.
To improve an agent's generalization ability to unseen environments, a speaker-follower model~\cite{speakerfollower} synthesizes new instructions for data augmentation, and ``environmental dropout'' method ~\cite{envdrop} is proposed to mimic unseen environments during training.
Wang~\etal~\cite{rcm} propose a reinforced cross-modal matching framework that combines the strength of reinforcement learning (RL) and imitation learning (IL) for vision language navigation task.
To estimate progress made towards the goal, Ma~\etal\cite{selfmonitor} introduce a self-monitoring method, which consists of visual-textual co-grounding module and a progress monitor. 
Hong~\etal\cite{graph} propose a language and visual entity relation graph that exploit the inter and intra modality relationships among the scene. 
Recently, VLBERT-based methods significantly improve performance on VLN tasks. Hong~\etal\cite{recurrent} develop a recurrent model that reuses the [CLS] token to maintain the history information. Qi~\etal~\cite{orist} propose an object-informed sequential BERT to encode visual perceptions and linguistic instructions.

\begin{figure*}[!t]
	\centering
	\includegraphics[width=1\linewidth]{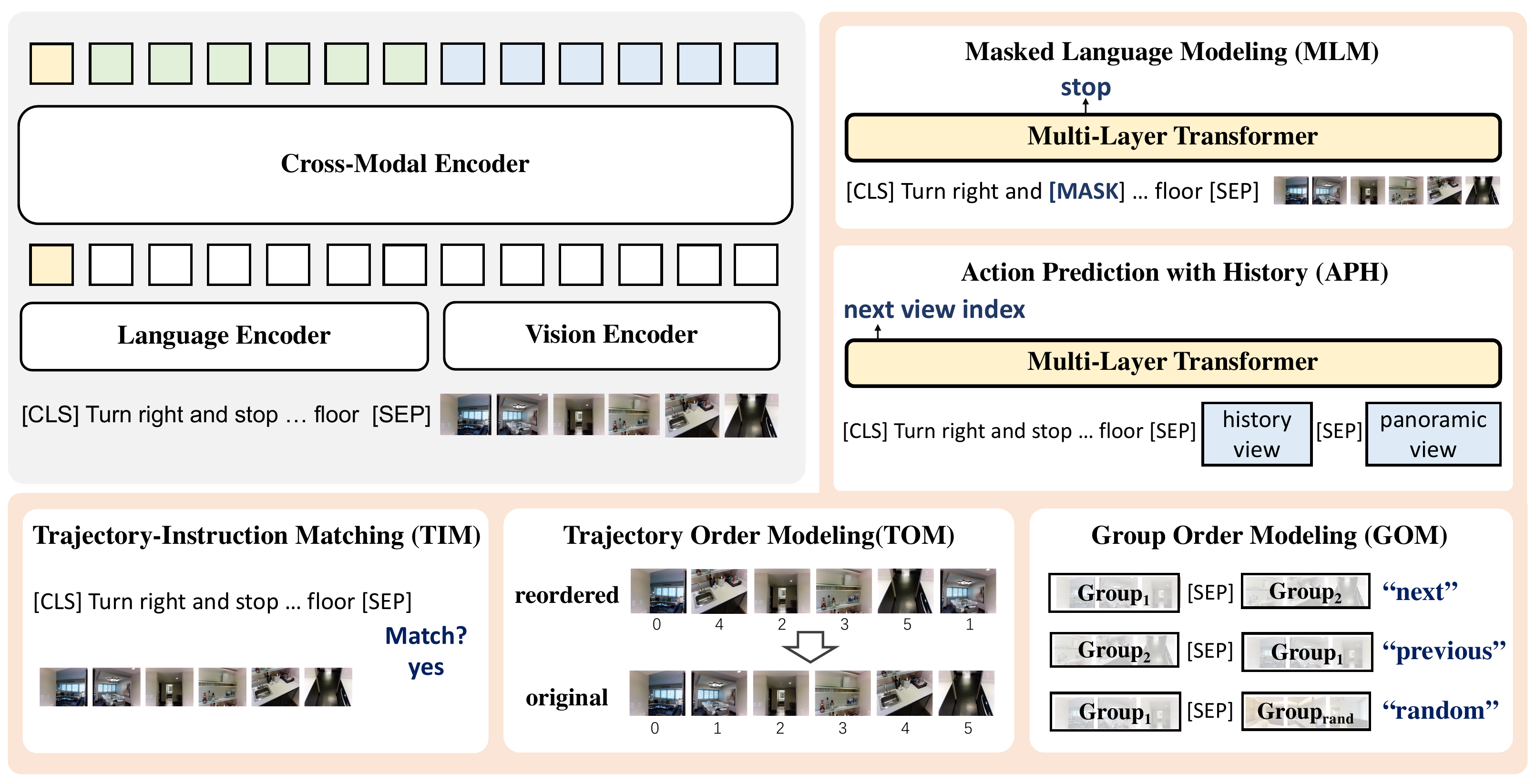}
	\vspace{-8mm}
	\caption{
	The main architecture of our pre-training model and five proxy tasks.
}
	\vspace{-6mm}
	\label{fig:architecture}
\end{figure*}

\vspace{-10pt}
\paragraph{Vision-Language Pre-training}
In recent years, many vision-language pre-training works~\cite{vilbert,lxmert,uniter,oscar} have been proposed to learn cross-modal representations for various vision language problems, such as Image-Text Retrieval~\cite{itr}, Referring Expression Grounding~\cite{reg} and Visual Question Answering~\cite{vqa}. 
Unlike conventional vision-language pre-training, VLN tasks additionally require the learned representations to facilitate action decision making. 
VLN-BERT~\cite{bertvln} performs path selection by predicting instruction and trajectory compatibility.
Similarly, AirBERT~\cite{airbert} trains the path-instruction matching task by collecting large numbers of indoor image-caption pairs. 
However, these methods ignore the importance of dynamic action decision. 
Although PREVALENT~\cite{prevalent} adds the task of action prediction (AP), the global-level path information cannot be considered because its input is a global instruction and a panoramic image of the current location,
which discards temporal visual context for action prediction.
{Moreover, all these methods do not explicitly mine temporal order information from either instructions or visual observations, which are crucial for an agent to predict actions.}
{In contrast, our pre-training is designed to relieve the above mentioned limitations, by introducing a \textit{history-aware} proxy task and two \textit{order-aware} proxy tasks. These tasks help the agent understand the history context and temporal order, facilitating the final action prediction.} We note ALTR~\cite{altr} introduces a ``Next Visual Scene'' task to capture the temporal context, but instead of predicting image orders it predicts visual features of future steps directly, which is a much harder task.

\section{Method}
\label{sec:HOP model}
\vspace{-3pt}
In this section, we first present the preliminaries of VLN to put our method in a proper context. Then, we describe the model architecture we adopted. Next, we provide the details of the five pre-training proxy tasks utilized in our proposed history-and-order aware pre-training paradigm. At last, we introduce the datasets used for pre-training.

\vspace{-2pt}
In VLN, a robot agent is randomly positioned in a 3D simulator with the initial state 
$\langle {u}_{0},\alpha_{0},\beta_{0} \rangle$, where ${u}_{0}$ is the starting viewpoint on the pre-defined navigation-graph, $\alpha_{0}$ and $\beta_{0}$ are angles of heading and elevation.
Given a natural language instruction $x = \langle x_1, x_2, \dots x_L \rangle$, where $L$ is the instruction length,
the agent needs to predict the next navigation action according to panoramic perceptions. Following the common practice, the action is predicted by selecting a navigable location from a candidate set provided by the simulator. Each navigable location is represented by its RGB image feature and its orientation feature. 

\subsection{Model Architecture}
The model architecture is illustrated on the top-left of Figure~\ref{fig:architecture}, which is similar to LXMERT~\cite{lxmert}. 
Taking the instruction-trajectory pair as input, the model first utilizes a language encoder and a vision encoder to extract single-modal representations from the instruction and image sequence, respectively. 
Then, these representations are fed into a cross-modal encoder to implement interactions between the two modalities and generate the final fused representations. 

\vspace{-13pt}
\paragraph{Language Encoder} 
We first use WordPieces~\cite{Wordpieces} to tokenize all words in an instruction, obtaining a sequence of tokens: $ \texttt{[CLS]}, w_1, w_2,\dots,w_L, \texttt{[SEP]}$, where $\texttt{[CLS]}$ and $\texttt{[SEP]}$ are added special tokens. 
Then, the text embedding of each token is obtained via summing up the token embedding and the position embedding, followed by Layer Normalization (LN). At last, the text embedding is passed through the single-modal language encoder, of which each layer consists of a self-attention sub-layer and a feed-forward sub-layer. The outputs of the language encoder are used as language features.

\vspace{-8pt}
\paragraph{Vision Encoder}
Trajectory $\tau = \langle v_1, v_2, \dots v_T \rangle$ represents the image sequences observed by the agent when traversing the environment, where $v_i$ is the observed image of the environment at step $i$ and $T$ is the number of total steps. To better capture order information from the trajectory, we use the front view image of the agent's observation at each position, rather than using the panoramic image. 
This is because panoramic images of the adjacent observation points in the same room are similar, causing difficulties for the agent to explore the dynamic and temporal information of the entire trajectory.

We first use ResNet-152~\cite{resnet} pre-trained on ImageNet~\cite{imagenet} to extract a 2048-dimensional image feature vector $v_{vis}$ for each front view image $v_i$. 
Then, we compute the orientation feature of heading $\alpha$ and elevation $\beta$ as $[\sin \alpha ; \cos \alpha ; \sin \beta; \cos \beta]$, and repeat it for 32 times to constitute a 128-dimensional direction feature vector $v_d$ as same as~\cite{envdrop}.
Each image $v_i$ in the trajectory is finally represented by a 2176-dimensional feature vector $ v_i = [v_{vis}; v_d]$ by concatenating $v_{vis}$ and $v_d$. 
At last, the image features of trajectory $\tau$ are passed through the single-modal vision encoder, of which each layer consists of a self-attention sub-layer and a feed-forward sub-layer. The outputs of the vision encoder are used as vision features.

\vspace{-8pt}
\paragraph{Cross-Modal Encoder}
We use the Cross-Modal Encoder to fuse features from both language and vision modalities. 
For the cross-modal encoder, each layer contains two self-attention sub-layers, one bi-directional cross-attention sub-layer and two feed-forward sub-layers. 
The outputs of the cross-modal encoder are used as cross-modal features for pre-training and downstream tasks.

Following~\cite{prevalent}, we set the layers' number $N_{text}$, $N_{image}$, $N_{cross}$ of text encoder, vision encoder and cross-modal encoder to 9, 1 and 3, respectively. 

\subsection{Pre-training Tasks}
\paragraph{Masked Language Modeling (MLM)} 
MLM is the most commonly used proxy task for BERT-based pre-training. For VLN pre-training, the goal of MLM is to recover masked words ${w_{m}}$ via reasoning over the surrounding words ${w_{\setminus m}}$ and the trajectory $\tau$.
Specifically, the inputs for MLM are the instruction $ w = \langle w_1, w_2,\dots,w_L \rangle$ and the corresponding trajectory $\tau = \langle v_1, v_2,\dots,v_T \rangle$. 
We randomly mask out the input tokens of the instruction with a 15\% probability, and replace the masked token ${w_m}$ with a special token \texttt{[mask]}.
This task is optimized by minimizing the negative log-likelihood:
\begin{equation}
    \mathcal{L}_{\text{MLM}}(\theta) = -\mathbb{E}_{({w}, {\tau})\sim D}\log P_{\theta}({w}_{m} | {w}_{\setminus {m}}, {\tau})\,,
\end{equation}
where $\theta$ denotes trainable parameters. 
Each pair $({w}, {\tau})$ is sampled from the training set $D$.

\vspace{-8pt}
\paragraph{Trajectory-Instruction Matching (TIM)} TIM is a global matching task, which is designed to predict whether a given image trajectory and an instruction are a matched pair.
The inputs for TIM are the instruction-trajectory pairs $\left(w,\tau \right)$.
During training, we generate negative samples by randomly replacing the trajectory with a mis-matched one, with a probability of 50\%. 
Specifically, the generated negative samples are selected only from the same environment, so that the model could focus on distinguishing between paths rather than environments. We use the output representation of the special token of \texttt{[CLS]} as the joint representation of the instruction-trajectory pair, and then feed it into an FC layer with a sigmoid function to predict the matching score $s_\theta(w,\tau)$.
We optimize this task via the binary cross-entropy loss:
\begin{equation}
    \mathcal{L}_{\text{TIM}}(\theta) = - \mathbb{E}_{({w}, {\tau})\sim D} [y \log P_{\theta} + (1-y) \log P_{\theta}],
\end{equation}
where $P_{\theta} = s_{\theta}({w}, {\tau})$, and $y\in \{0, 1\}$ indicates whether the sampled trajectory-instruction pair is a match. 

\vspace{-8pt}
\paragraph{Trajectory Order Modeling (TOM)}
VLN is sensitive to the sequence order of trajectory,
thus we design the TOM task to enable the model to learn the temporal order within instructions in addition to visual-textual correspondence.
The inputs for TOM are the instruction $w$ and the reordered trajectory ${\tau}'$. 
Specifically, we randomly selected 50\% images of the original trajectory $\tau$ for shuffling. 
The goal of TOM is to reconstruct the correct order ${r}={\langle}{r_1, r_2, ..., r_N}{\rangle}$ of the original trajectory $\tau$ 
with reference to the given instruction $w$, where $N$ is the number of steps of trajectory.
This task is formulated as a classification problem of $N$ classes. We feed the vision output of the cross-modal encoder into an FC layer with softmax to predict the order $r'_{k}$ for each image $k$ in the reordered trajectory ${\tau}'$, by minimizing the cross-entropy loss:

\vspace{-5pt}
\begin{equation}
    \mathcal{L}_{\text{TOM}}(\theta) = -\mathbb{E}_{({w}, {\tau}')\sim D} \sum_{i=1}^{N}y_i\log P_{\theta}(r'_k | {w}, {\tau}')\,,
\end{equation}
where $y_{i}=1$ if the predicted order $r'_k$ for image $k$ is the original order $i$, otherwise $y_{i}=0$.

\begin{figure}[!t]
	\begin{center}
		\includegraphics[width=1.0\linewidth]{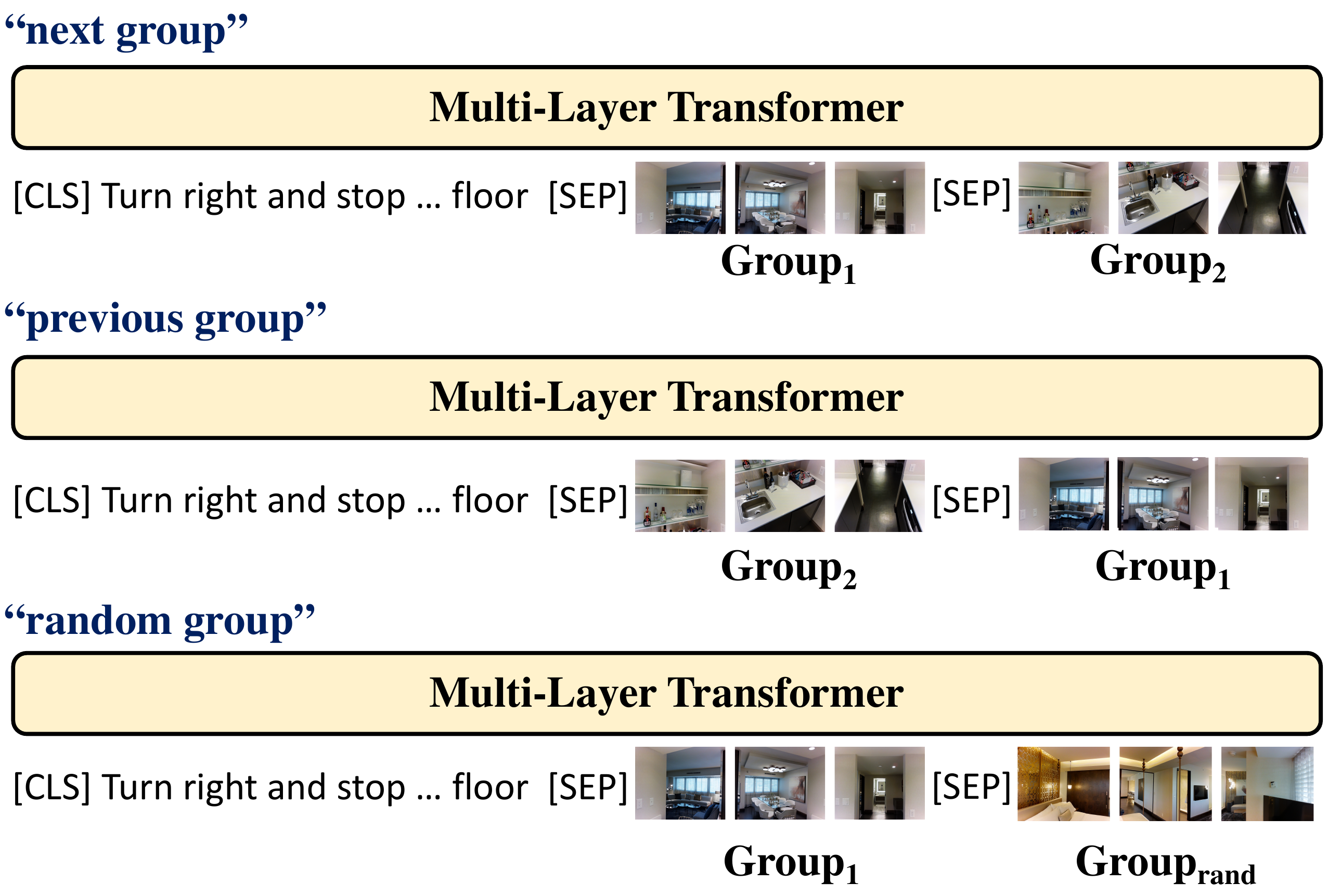}
	\end{center}
	\vspace{-13pt}
	\caption{Illustration of Group Order Modeling (GOM).
	}
	\label{fig:NVP}
	\vspace{-13pt}
\end{figure}

\vspace{-8pt}
\paragraph{Group Order Modeling (GOM)} 
This task shares the same motivation as TOM but at a sub-trajectory level. It  predicts previous, next, or random relations between two sub-trajectories. 
As shown in Figure~\ref{fig:NVP}, the inputs for GOM are the instruction $w$ and image sequence group ($G_1, G_2$) that is derived from the trajectory ${\tau}$. 
Specifically, we divide the trajectory into two parts ($G_1, G_2$) sequentially in an even way.
Furthermore, there is a probability of $1/3$ that $G_2$ will be placed after $G_1$,  a probability of $1/3$ that $G_2$ will be placed before $G_1$, and a remaining  probability of $1/3$ that $G_2$ will be replaced by random sampling from image sequence groups of other trajectories. 
This task is cast as a classification problem of three classes.
If $G_1$ happens before $G_2$, we denote it as $c=1$; if $G_1$ happens after $G_2$, we denote it as $c=2$; if $G_2$ is an image sequence group randomly sampled from different environments, we denote it as $c=3$. 
The special token \texttt{[SEP]} is used to separate the two groups.
We use the representation of \texttt{[CLS]} token as the joint embedding of the input visual and textual information. Then we apply an FC layer with softmax  to make a three-class prediction of $c'$. This task is optimized by minimizing the cross-entropy loss:
{\small
\begin{equation}
    \mathcal{L}_{\text{GOM}}(\theta) = -\mathbb{E}_{({w}, {(G1,G2)})\sim D} \sum_{c}y_c\log P_{\theta}(c'| {w}, {(G1,G2)}),
\end{equation}
}%
where $y_c\in \{0, 1\}$ indicates whether the predicted class $c'$ is the desired class $c$ or not.

\vspace{-5pt}
\paragraph{Action Prediction with History (APH)}
The motivation of this task is to make the learned representation benefiting the final goal: predicting navigation action. 
The inputs for APH are the instruction $w$, the history trajectory $\tau_{t-1} =\langle v_{1}, v_{2}, \dots, v_{t-1}\rangle$, and the panoramic view $\textit{\textbf{p}} = \{ p^1, p^2, \dots, p^{36}\}$ of the current step $t$.
The panoramic view consists of 36 images from 12 surrounding angles, each with 3 camera poses (up, down, horizon).
As in PREVALENT~\cite{prevalent}, action decision is achieved by selecting next view image $v'_{t+1}$ from the candidate views (\ie panorama observation $v_{t_p}$), which can be expressed as a classification problem.
The output on the special token \texttt{[CLS]} represents a fused representation of the two modalities. We apply an FC layer to the representation of \texttt{[CLS]} to predict the next view $v'_{t+1}$.
We optimize this task via a cross-entropy loss:
{\small
\begin{equation}
    \mathcal{L}_{\text{APH}}(\theta) = -\mathbb{E}_{({w}, {\tau},{v_{pano}})\sim D}\sum_{p}{y_p\log P_{\theta}(v'_{t+1}|w, {\tau_{t-1}}, v_{t_p}}),%
\end{equation}%
}%
where $p$ represents labels of the 36 images in the panoramic view image, and $y_p\in \{0, 1\}$ indicates whether the predicted next view image $v'_{t+1}$ is the desired next view image of label $p$ or not.

\subsection{Pre-training Datasets}
We construct our pre-training dataset based on existing datasets: PREVALENT~\cite{prevalent} and BnB~\cite{airbert}. PREVALENT uses a pre-trained speaker model to produce more instructions to augment R2R dataset. It contains 104K original R2R samples and 6482K synthesized samples. BnB dataset collects image-caption pairs from Airbnb. We use raw images and captions from the BnB dataset and reprocessed them. 
Indeed, nearly half of the BnB images are captionless (\ie images without captions).
Thus, to better adapt BnB dataset to the designed pre-training tasks such as Trajectory Order Modeling, we remove these captionless images. 
To construct path-instruction pairs, we concatenate the images and concatenate the corresponding captions. 
Each path contains 5-7 images, which is consistent with the R2R dataset. 
For image features, we used a Resnet-152 network pre-trained on ImageNet to extract a mean-pooled feature vector, same as the encoding method of images in Matterport3D.
Our processed BnB data contains 342K image sequence-caption pairs. 

\section{Experiments}
\label{sec:experiments}
In this section, we conduct comprehensive experiments on several downstream VLN tasks and provide detailed ablation studies to validate the effectiveness of our proposed method. 
\subsection{Downstream tasks}
We focus on four downstream VLN tasks that are all based on the Matterport3D simulator~\cite{r2r}: Room-to-Room (R2R)~\cite{r2r}, Room-across-Room (RxR)~\cite{rxr}, Navigation from Dialog History (NDH)~\cite{ndh} and REVERIE~\cite{reverie}. R2R serves as an in-domain task, and the other three serve as out-of-domain tasks. These tasks have different characteristics and evaluate the agent from different views.
\begin{itemize}[itemsep=2pt,topsep=2pt,parsep=2pt]
    \item R2R and RxR are VLN tasks with low-level, fine-grained instructions that aim at verifying the agent's ability to generalize to unseen environments.
    \item REVERIE is a VLN task with high-level instructions, focusing on grounding remote target object.
    \item NDH is a VLN task that uses indirect instructions such as dialog history, which can be used  to study the agent's generalization ability to new tasks.
\end{itemize}

\begin{table*}[!t]
\centering
\resizebox{0.85\linewidth}{!}{
\begin{tabular}{@{\hspace{3pt}}l @{}c@{\hspace{9pt}}c@{\hspace{9pt}}c@{\hspace{9pt}}c|c@{\hspace{9pt}}c@{\hspace{9pt}}c@{\hspace{9pt}}c|c@{\hspace{9pt}}c@{\hspace{9pt}}c@{\hspace{9pt}}c}
\toprule
\multicolumn{1}{l}{\multirow{2}{*}{Methods}}& \multicolumn{4}{c}{R2R Validation Seen} & \multicolumn{4}{c}{R2R Validation Unseen} & \multicolumn{4}{c}{R2R Test Unseen} \\ 
~  & TL & NE $\downarrow$ & SR $\uparrow$ & SPL $\uparrow$ & TL & NE $\downarrow$ & SR $\uparrow$ & SPL$\uparrow$  & TL & NE $\downarrow$ & SR $\uparrow$ & SPL $\uparrow$\\ 
\midrule
SF~\cite{speakerfollower} & - & 3.36 & 66 & - & - & 6.62 & 35 & - & 14.82 & 6.62 & 35 & 28\\
RCM~\cite{rcm} & 10.65 & 3.53 & 67 & - & 11.46  & 6.09 & 43 & - & 11.97 & 6.12 & 43 & 38 \\
Regretful~\cite{regretful}& - & 3.23 & 69 & 63 & - & 5.32 & 50 & 41 & 13.69 & 5.69 & 48 & 40 \\
Fast-short~\cite{fast} & - & - & - & - & 21.17 & 4.97 & 56 & 43 & 22.08 & 5.14 & 54 & 41 \\
EnvDrop~\cite{envdrop} & 11.00 & 3.99 & 62 & 59 & 10.70 & 5.22 & 52 & 48 & 11.66 & 5.23 & 51 & 47 \\
OAAM~\cite{oaam} & 10.20 & - & 65 & 62 & 9.95 & - & 54 & 50 & 10.40 & 5.30 & 53 & 50 \\
EntityGraph~\cite{graph} & 10.13 & 3.47 & 67 & 65 & 9.99 & 4.73 & 57 & 53 & 10.29 & 4.75 & 55 & 52\\
NvEM~\cite{nvem} & 11.09 & 3.44& 69& 65 &11.83 &4.27& 60& 55 &12.98 &4.37 &58& 54\\
ActiveVLN\cite{avig} & 19.70 & 3.20 & 70 & 52 & 20.6 & 4.36 & 58 & 40 &  21.6 & 4.33 & 60 & 41\\
\hline
Press~\cite{press}  & 10.57 & 4.39 & 58 & 55 & 10.36 & 5.28 & 49 & 45 & 10.77 & 5.49 & 49 & 45 \\
PREVALENT~\cite{prevalent} & 10.32 & 3.67 & 69 & 65 & 10.19 & 4.71 & 58 & 53 & 10.51 & 5.30 & 54 & 51\\
RecBERT~\cite{recurrent} & 11.13 & 2.90 & 72 & 68 & 12.01 & 3.93 & \textbf{63} & \high{57} & 12.35 & 4.09 & \textbf{63} & 57\\
AirBERT~\cite{airbert} & 11.09 & \textbf{2.68} & 75 & \high{70} & 11.78 & 4.01 & 62 & \textbf{56} & 12.41 & 4.13 & 62 & 57 \\
\midrule
HOP ($0$) & 10.75 & 3.50 & 66 & 63 & 11.80 & 4.74 & 54 & 49 & 12.53 & 4.93 & 55 & 50 \\
HOP ($1$) & 11.51 & \high{2.46} & \high{76} & \high{70} & 12.52 & \high{3.79} & \high{64} & \high{57} & 13.29 & \textbf{3.87} & \high{64} & \textbf{58}\\
HOP ($2$) & 11.26 & 2.72 & \textbf{75} & \high{70} & 12.27 & \textbf{3.80} & \high{64} & \high{57} & 12.68 & \high{3.83} & \high{64} & \high{59}\\
\bottomrule
\end{tabular}}
\vspace{-1mm}
\caption{
Comparison with state-of-the-art methods on R2R.
First group are no pre-training methods. The second group are existing pre-training-based methods. The third group are our methods.
HOP ($0$) denotes our baseline model without pre-training. HOP ($1$) denotes finetuned model pre-trained on the same data as PREVALENT. HOP ($2$) denotes finetuned model pre-trained on data of both PREVALENT and our processed data from BnB.
\high{Blue}  and \textbf{Black} denote the best and runner-up results, respectively.}
\vspace{-2mm}
\label{tab:main_result_r2r}
\end{table*}

\subsection{Implementation Details}
\paragraph{Pre-training}
We use 4 Tesla V100 GPUs for pre-training. The batch size for each GPU is set to 128. AdamW~\cite{adamw} optimizer is adopted and the learning rate is set to $5\!\times\!10^{-5}$. The model is trained for 15 epochs. 
We conduct task sampling training for each mini-batch. 
For each mini-batch, we  choose only one of the five proxy tasks to train the model.
\vspace{-10pt}

\paragraph{Fine-tuning}
Different from PREVALENT~\cite{prevalent} that only uses the pretrained language representations to finetune downstream tasks, 
we use an architecture similar to RecBERT~\cite{bertvln} as our baseline for finetune, of which both the image and language representations could be used for downstream tasks.
Following RecBERT, we employ a recurrent function to update state \texttt{[CLS]} and use its attention distribution on navigation candidates to determine the next action. For more details please refer to~\cite{recurrent}.
For R2R task, we set the batch size to 16 and the learning rate to $1\!\times\!10^{-5}$. 
Following previous works~\cite{recurrent}, we use both the original training data of R2R and the augmented data from PREVALENT~\cite{prevalent} to train the agent.
For both NDH and REVERIE tasks, we set the batch size to 8 and the learning rate to $1\!\times\!10^{-5}$.
All the above three downstream tasks are finetuned on a single 1080Ti GPU. 
For RxR task which has longer instructions, we finetune on a single V100 GPU, and we set the batch size to 16 and the learning rate to $7\!\times\!10^{-6}$.

\subsection{Results}
For each of the four downstream VLN tasks, we first introduce the evaluation metrics used for the task, and then we compare our method with SoTA methods. 
Specifically, we report results of our method on three settings: (I) baseline results without pre-training, which is denoted as HOP (0); (II) finetuned results with pre-training only using data of PREVALENT, which is denoted as HOP (1); (III) finetuned results with pre-training using data of PREVALENT and processed data from BnB, which is denoted as HOP (2).

\vspace{-8pt}
\subsubsection{Room-to-Room (R2R)}

\vspace{-5pt}
\paragraph{Evaluation Metrics}
Four commonly used metrics are adopted:
Trajectory Length (TL) that measures the average length of the navigation trajectory in meters;
Navigation Error (NE) which is the mean of the shortest path distance in meters between the agent’s stop location and the target location; 
Success Rate (SR) that measures the ratio of successful tasks, of which the agent's stop position is less than 3 meters away from the target position; 
Success rate weighted by Path Length (SPL)~\cite{spl} that measures both the accuracy and efficiency of navigation. SPL is the key metric for R2R.

\begin{table}[!t]
\centering
\resizebox{\linewidth}{!}{
\begin{tabular}{@{\hspace{3pt}}l@{}r@{\hspace{9pt}}c@{\hspace{9pt}}c|r@{\hspace{9pt}}c@{\hspace{9pt}}c r@{\hspace{9pt}}c@{\hspace{9pt}}c }
\toprule
\multicolumn{1}{l}{\multirow{2}{*}{Methods}} & \multicolumn{3}{c}{NDH Validation Unseen} & \multicolumn{3}{c}{NDH Test Unseen} \\ 
~  & \textbf{Oracle}  & \textbf{Navigator} & \textbf{Mixed}  & \textbf{Oracle} & \textbf{Navigator} & \textbf{Mixed}\\ 
\midrule
Seq2Seq~\cite{ndh}  & 1.23 & 1.98 & 2.10 & 1.25 & 2.11 & 2.35  \\
CMN~\cite{cmn} & 2.68 & 2.28 & 2.97 & 2.69 & 2.26 & 2.95\\
PREVALENT~\cite{prevalent}  & 2.58  & 2.99 &  3.15 &
1.67 & 2.39& 2.44 \\
ORIST~\cite{orist} & 3.30 & 3.29 & 3.55 & 2.78 & 3.17 & 3.15 \\
\midrule
HOP (0)& 3.08& 3.10 & 3.38 & 2.05 & 2.12 & 2.26 \\
HOP (1)& \textbf{3.96} & \textbf{3.99} & \textbf{4.37} & \textbf{2.92} & \high{3.20}& \high{3.31}\\
HOP (2)& \high{4.07} &\high{4.05} & \high{4.41} & \high{2.99} & \textbf{3.18}&\textbf{3.24}\\
\bottomrule
\end{tabular}}
\vspace{-1mm}
\caption{Comparison with   state-of-the-art methods on NDH measured by the Goal Progress (m). 
}
\label{tab:main_result_cvdn}
\vspace{-6mm}
\end{table}

\begin{table*}[!t]
\centering
\resizebox{\linewidth}{!}{
\begin{tabular}{l cccccc|cccccc|cccccc}
\toprule
\multicolumn{1}{l}{\multirow{3}{*}{Methods}} & \multicolumn{6}{c}{REVERIE Validation Seen} &\multicolumn{6}{c}{REVERIE Validation Unseen} & \multicolumn{6}{c}{REVERIE Test Unseen} \Tstrut\\
\cline{2-19}
~& \multicolumn{4}{c}{Navigation}  &  \multicolumn{1}{c}{\multirow{2}{*}{RGS$\uparrow$}}&
\multicolumn{1}{c|}{\multirow{2}{*}{RGSPL$\uparrow$}} & \multicolumn{4}{c}{Navigation}  & \multicolumn{1}{c}{\multirow{2}{*}{RGS$\uparrow$}}&
\multicolumn{1}{c|}{\multirow{2}{*}{RGSPL$\uparrow$}} & \multicolumn{4}{c}{Navigation}   & \multicolumn{1}{c}{\multirow{2}{*}{RGS$\uparrow$}}&
\multicolumn{1}{c}{\multirow{2}{*}{RGSPL$\uparrow$}}  \\
~& \multicolumn{1}{c}{SR$\uparrow$} & \multicolumn{1}{c}{OSR$\uparrow$} & \multicolumn{1}{c}{SPL$\uparrow$}  & \multicolumn{1}{c}{TL} &  &  & \multicolumn{1}{c}{SR$\uparrow$} & \multicolumn{1}{c}{OSR$\uparrow$} & \multicolumn{1}{c}{SPL$\uparrow$} &\multicolumn{1}{c}{TL} & & & \multicolumn{1}{c}{SR$\uparrow$} & \multicolumn{1}{c}{OSR$\uparrow$} & \multicolumn{1}{c}{SPL$\uparrow$} & \multicolumn{1}{c}{TL} &  & \Tstrut\\

\hline
Human & -- & --  & -- & {--} & -- & -- & -- & --  & --  & {--} & -- & -- & 81.51 & 86.83  & 53.66 &{21.18} & 77.84 &  51.44 \\
\hline
RCM \cite{rcm} & 23.33 & 29.44 & 21.82& 10.70 & 16.23 & 15.36 & 9.29 & 14.23 & 6.97 &11.98 & 4.89 & 3.89 & 7.84 & 11.68 & 6.67 & 10.60 & 3.67 & 3.14\\
SMNA \cite{selfmonitor} & 41.25& 43.29  & 39.61& 7.54   & 30.07 &  28.98 & 8.15 & 11.28 &6.44 &9.07 & 4.54& 3.61 & 5.80& 8.39 & 4.53 & 9.23  & 3.10& 2.39 \\
FAST-Short \cite{fast} & 45.12& 49.68 &40.18&13.22  &31.41 & 28.11 & 10.08 & 20.48 & 6.17 &29.70 & 6.24 & 3.97 & 14.18 & 23.36 & 8.74 & 30.69  & 7.07 & 4.52 \\
FAST-MATTN \cite{reverie} &50.53 & 55.17 & 45.50 & 16.35&31.97 & 29.66 & 14.40 & 28.20 & 7.19 & 45.28  & 7.84 & 4.67 & 19.88 & {30.63} & 11.61 & 39.05 & 11.28 & 6.08 \\
ORIST~\cite{orist} & 45.19 & 49.12 & 42.21 & 10.73 & 29.87 & 27.77 & 16.84 & 25.02 & 15.14 & 10.90 & 8.52 & 7.58 & 22.19 & 29.20 & 18.97 & 11.38 & 10.68 & 9.28\\
RecBERT~\cite{recurrent} & 51.79 & 53.90 & 47.96& 13.44 & 38.23 & \textbf{35.61} &30.67 & 35.02 & 24.90 & 16.78 & \textbf{18.77} & 15.27 & 29.61 & 32.91 & 23.99 & 15.86 & 16.50 & 13.51 \\
AirBERT~\cite{airbert}& 47.01 & 48.98 & 42.34 & 15.16 & 32.75 & 30.01 & 27.89 & 34.51 & 21.88 & 18.71 & 18.23 & 14.18 & 30.28 & 34.20 & \textbf{23.61} & 17.91 & 16.83 & 13.28 \\
\hline
HOP (0) & 43.78 & 46.03 & 40.11 & 11.67 & 28.95 & 26.69 & 24.17 & 30.16 & 20.07 & 16.52 & 12.35 & 10.18 & 23.12 & 26.27 & 18.5 & 16.15 & 11.17 & 9.1\\
HOP (1)&54.81& 56.08& \high{48.05}& 14.05 & \high{40.55} & \high{35.79} & 30.39 & 35.30& \textbf{25.10}& 17.16 & 18.23 & \textbf{15.31} & 29.12 & 32.26 & 23.37 & 17.05 & \textbf{17.13} & \textbf{13.90} \\
HOP (2)& 53.76 & 54.88 & \textbf{47.19} & 13.80 & \textbf{38.65} & 33.85 & 31.78 & 36.24 & \high{26.11} & 16.46 & \high{18.85} & \high{15.73} & 30.17 & 33.06 & \high{24.34} & 16.38 & \high{17.69} & \high{14.34}\\
\bottomrule
\end{tabular}}
\vspace{-1mm}
\caption{Comparison with the state-of-the-art methods on REVERIE. SPL is the main metric for its navigation sub-task, and RGSPL is the main metric for the REVERIE task.}
\label{tab:reverie}
\vspace{-1mm}
\end{table*}

\vspace{-10pt}
\paragraph{Comparison with SoTA}
Table~\ref{tab:main_result_r2r} presents the results on R2R task. It shows that our model outperforms other methods on all data splits and metrics. 
Specifically, our method achieves better performance than other SoTA methods, such as RecBERT and AirBERT. Our method outperforms them with a large margin of 2\% in terms of the main metric SPL  on the test unseen split. 
Note that RecBERT is initialised from pre-trained model of PREVALENT and shares the same architecture as ours.
This indicates that our pre-training can effectively improve the navigation ability of the agent.
If removing our pre-training, we observe significant performance drop  on  all metrics, as shown by the results of HOP (0). In particular, it drops 9\% of SR and 9\% of SPL on the test unseen split.
\vspace{-8pt}
\subsubsection{Navigation from Dialog History (NDH)}

\paragraph{Evaluation Metric}
NDH evaluates the performance using Goal Progress (GP) in meters, which measures the average progress of the agent towards the target location. 
\vspace{-10pt}
\paragraph{Comparison with SoTA}
The results are shown in Table~\ref{tab:main_result_cvdn}. Our model outperforms the SoTA method ORIST \cite{orist} on both validation and test unseen environments in all settings. Particularly, our HOP (2) achieves up to 1 meter improvement over ORIST on validation unseen split under the mixed setting. Our method also performs much better than the pre-training method PREVALENT (about 1.3 meter improvement on the Test split under the Oracle setting). These results demonstrate the effectiveness and generalization ability of our pre-trained model.

\vspace{-0pt}
\subsubsection{REVERIE}

\paragraph{Evaluation Metrics}
REVERIE uses the same metrics as R2R to evaluate its navigation sub-task.
Additionally, Oracle Success Rate (OSR), Remote Grounding Success rate (RGS) and RGS weighted by Path Length (RGSPL) are adopted. 
OSR measures the ratio of tasks of which one of its trajectory viewpoints can observe the target object within 3 meters. RGS measures the ratio of tasks that successfully locate the target object.
RGSPL is RGS weighted by Path Length, which is the main metric for this task.

\vspace{-0pt}
\paragraph{Comparison with SoTA}
The results are presented in Table~\ref{tab:reverie}. 
It shows that our model outperforms previous methods  on all the splits in terms of the main metric SPL for navigation sub-task and in terms of the main metric RGSPL for the entire REVERIE task.
In addition, we also note that on the Test split, although our method performs slightly ($\downarrow0.1$) worse than AirBERT according to SR for navigation, our SPL and RGSPL results are significantly better ($\uparrow1.1$) than AirBERT. This indicates our method is more efficient on navigation and object grounding. 


\subsubsection{Room-Across-Room (RxR)}

\paragraph{Evaluation Metrics}
In addition to the aforementioned SR and SPL metrics, RxR also adopts normalized Dynamic Time Warping (nDTW)~\cite{ndtw} and success rate weighted by Dynamic Time Warping (sDTW) for evaluating performance. These two metrics are designed to measure the path fidelity compared to the ground truth path. 

\vspace{-0pt}
\paragraph{Comparison with SoTA}

As shown in Table~\ref{tab:main_result_rxr}, our model performs favourably against SoTA methods in terms of all metrics. 
In particular, our model achieves 3.1\% improvement in SR over previous SoTA  on the unseen split.

\begin{table}[!t]
\centering
\resizebox{\linewidth}{!}{
\begin{tabular}{l cccc|cccc}
\toprule
\multicolumn{1}{l}{\multirow{2}{*}{Methods}} & \multicolumn{4}{c}{RxR Validation Seen} & \multicolumn{4}{c}{RxR Validation Unseen} \\ 
~  & SR$\uparrow$ & SPL $\uparrow$ & nDTW $\uparrow$ & sDTW $\uparrow$ & SR$\uparrow$ & SPL $\uparrow$ & nDTW $\uparrow$ & sDTW$\uparrow$ \\ 
\midrule
Baseline~\cite{rxr}&28.6 & - & 0.45& 0.23  & 26.1&-& 0.42 & 0.21 \\
EnvDrop~\cite{envdrop}& 48.1 & 0.44 & 0.57 & \high{0.40} & 38.5 & 0.34 & 0.51 & 0.32\\
+Syntax~\cite{syntax}& 48.1 & 0.44 & \high{0.58} & \high{0.40} & 39.2 & 0.35 & \high{0.52} & 0.32\\
\midrule
HOP (0)& 42.0 & 0.41 & 0.51 & 0.34& 36.3 & 0.31 & 0.48 & 0.29\\
HOP (1)& 48.3& \high{0.45} & 0.57 & \high{0.40} & 42.1 & \high{0.36} & 0.51 & \high{0.33}  \\
HOP (2)& \high{49.4} & \high{0.45} & \high{0.58} & \high{0.40} & \high{42.3} & \high{0.36} & \high{0.52} & \high{0.33} \\
\bottomrule
\end{tabular}}
\vspace{-1mm}
\caption{Comparison with state-of-the-art methods on RxR using English instructions. 
}
\label{tab:main_result_rxr}
\vspace{-0mm}
\end{table}

\begin{table*}[!t]
\small
\centering
\resizebox{\linewidth}{!}{
\begin{tabular}{lcl  cc|ccccc|c}
\toprule
 \multicolumn{1}{l}{\multirow{2}{*}{Pre-training Data}}& &\multicolumn{1}{l}{\multirow{2}{*}{Pre-training Tasks}}
 & \multicolumn{2}{c}{R2R} & \multicolumn{5}{c}{REVERIE} &  \multicolumn{1}{c}{NDH}\\
& & &SR & SPL & SR & OSR & SPL & RGS & RGSPL & Goal Progress\\ 
\midrule
None & 1 & None & 54.19 & 49.35 & 24.17 & 30.16 & 20.07 & 12.35 & 10.18 & 3.38\\
\midrule
\multicolumn{1}{l}{\multirow{7}{*}{PREVALENT}}&2& MLM & 60.75 & 54.81 & 27.18 & 31.84 & 21.83 & 15.31 & 12.48 &3.76 \\
&3 & MLM + TIM & 61.52 & 55.31& 28.06 & 34.76 & 22.84 & 16.30 & 13.29& 3.86\\
&4 & MLM + TOM & 61.81 & 54.96& 28.57 & 31.27 & 22.67 & 17.98 & 14.49& 3.88\\
&5 & MLM + GOM & 61.98 & 55.22& 27.83 & 35.08 & 22.53 & 17.24 & 14.14& 3.84\\
&6 & MLM + APH &62.01& 56.13 & 29.37 & 34.88 & 23.76 & 17.52 & 14.16 & 3.88\\
&7 & MLM + AP  & 61.27 & 55.68& 28.69 & 33.20& 23.25& 16.76 & 13.64& 3.83\\

&8 & MLM + TIM + TOM + GOM & 63.09& 56.61 & 29.99 & 35.13 & 24.66 & 18.03 & 15.06 & 4.04\\
&9 & MLM + TIM + TOM + GOM + APH &\high{63.86} & \textbf{57.07} & \textbf{30.39} & \textbf{35.30} & \textbf{25.10} & \textbf{18.35} &\textbf{15.31}&\textbf{4.37}\\
\midrule
PREVALENT + BnB* &10 & MLM + TIM + TOM + GOM + APH & \textbf{63.52} & \high{57.22} & \high{31.78} & \high{36.24} & \high{26.11} & \high{18.85} & \high{15.73}&\high{4.41}\\
\bottomrule
\end{tabular}}
\vspace{-0mm}
\caption{Ablation study of the pre-training tasks and data. We use the task of R2R, REVERIE and NDH as benchmarks. Where BnB* represents our processed data from BnB dataset.}
\label{tab:ablation}
\vspace{-0mm}
\end{table*}

\subsection{Ablation Study}
\vspace{3pt}
\paragraph{The Effect of Pre-training Tasks}
To evaluate the effectiveness of different pre-training tasks, we conduct an ablation study on R2R, REVERIE and NDH validation unseen set. The results are presented in Table~\ref{tab:ablation}.

First, we evaluate the effect of the generic MLM task alone. Model 1 shows the results of the baseline model directly trained on downstream VLN tasks without any pre-training. Model 2 shows the results when only applying MLM for pre-training. MLM brings large improvements on all downstream VLN tasks, especially on the R2R task ($\uparrow 6\% $ SR). 

Secondly, we evaluate the effect of our proposed pre-training tasks that are specifically designed for VLN, by combining these tasks with MLM during pre-training. 
Model 3$\sim$ Model 6 present the results of combining MLM with the task of TIM, TOM, GOM, and APH respectively. The result shows that all these four proxy tasks can further improve the navigation performance.
Among these four tasks, APH contributes the most, then TOM, GOM, and TIM. 
This demonstrates that action prediction with history information indeed helps learn better representations.

Furthermore, we find that these proxy tasks are complementary. When we combine these tasks together, the combined scores (Model 8 and Model 9) are much higher than the separate scores (Model 3$\sim$ Model 6).
We observe a gain of 9\% SR on R2R, 6\% SR on REVERIE, 1 meter on NDH compared to the baseline model without pre-training.

\vspace{-3pt}
\paragraph{The Effect of Pre-training Data}
As shown in Model 10 of Table~\ref{tab:ablation}, the model pre-trained with data from both PREVALENT and BnB* (our processed BnB data) achieves the best performance. 
We find that the model achieves a significant improvement on REVERIE task when pre-trained with additional BnB* data than only with PREVALENT data, while keep competitive on R2R and NDH task.
This could be because BnB's captions primarily describe rooms and objects, which matches REVERIE's mission of grounding objects.

\begin{figure}[!t]
	\begin{center}
		\includegraphics[width=0.85\linewidth]{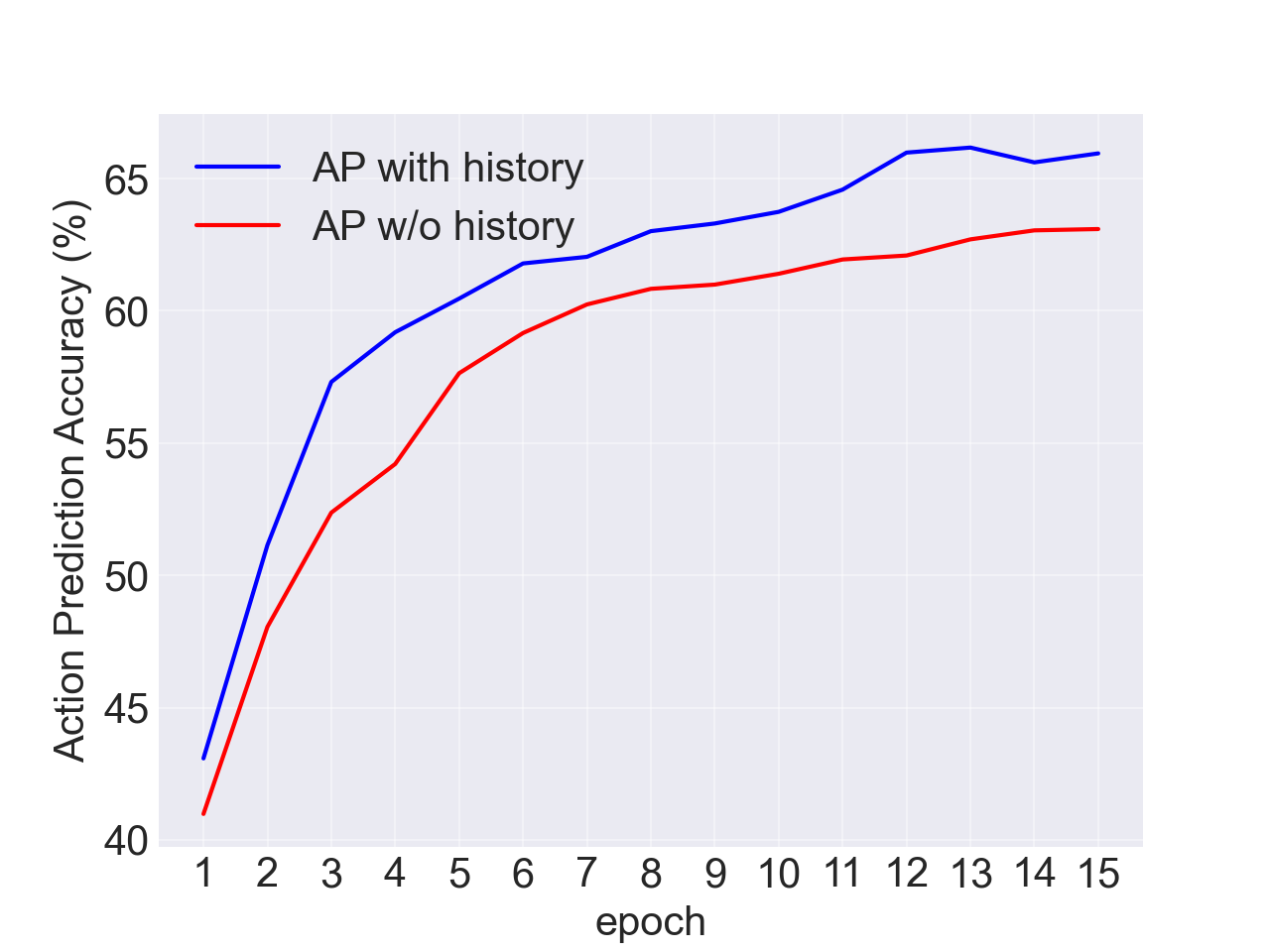}
	\end{center}
	\vspace{-10pt}
	\caption{APH \textit{v.s.} AP regarding action prediction accuracy.
	}
	\label{fig:AP}
	\vspace{-7pt}
\end{figure}

\vspace{-3pt}
\paragraph{The Effect of History Information in APH}
We also pre-train our model using the Action Prediction task without history to verify the importance of referring history information for decision making. 
As shown in the validation curves of Action Prediction of Figure~\ref{fig:AP}, our APH converges faster than AP during pre-training, and results in higher accuracy. 
In addition, as shown in Table~\ref{tab:ablation} (Model 6 and  Model 7), Action Prediction with History achieves better performance on all three downstream tasks and all metrics than without history. Therefore, the history information is beneficial to the vision-language pre-training for VLN tasks.

\subsection{Limitations and Future work}
Like most of existing pre-training methods, the training of our model also requires a large number of computational resources. In the future, we will explore more efficient model architectures. 
In addition, our current work focuses on indoor and navigation graph based environments. We can devote efforts on outdoor and/or continuous environments in the future.


\section{Conclusion}
In this paper, we present a history-and-order aware pre-training (HOP) paradigm to solve the issues existing in the previous Vision-and-Language Navigation pre-training methods.
We first carefully examine and compare previous methods, and we find these methods either overlook the important historical context in pre-training or neglect the role of the action ordering. Thus, we design an Action Prediction with History (APH) task that provides history visual observations to the action prediction in the pre-training. We then propose two order-aware proxy tasks, including Trajectory Order Modeling (TOM) and Group Order Modeling (GOM). 
These two tasks equip the pre-trained model with the ability to understand the temporal order within instructions, in addition to the widely used visual-textual matching capability.
Extensive experimental results on four mainstream downstream VLN tasks, including R2R, RxR, NDH and REVERIE, demonstrate the effectiveness of our proposed VLN pre-training paradigm, the HOP.

{\small
\bibliographystyle{ieee_fullname}
\bibliography{egbib}
}

\clearpage

\appendix
\section*{Appendices}
\section{Downstream tasks}
\subsection{Room-to-Room (R2R)}
The task of R2R~\cite{r2r} requires agents to follow detailed instructions to navigate from one room to another. These instructions contain rich linguistic information, such as ``Walk out of the room and down the hallway. Walk past the kitchen and stop outside of the door to the dining room''.
\vspace{0pt}

The R2R dataset is collected from Matterport3D simulator. It contains 21,567 navigation instructions, 7,189 trajectories and 10,800 panoramic views of 90 real-world building-scale indoor environments. The average length of each instruction is 29 words. The R2R dataset consists of four splits: train, validation seen and validation unseen, test unseen. 

\vspace{0pt}
\subsection{Room-Across-Room (RxR)}
The task of RxR~\cite{rxr} is an updated version of R2R and is more challenging than R2R. For example, instructions in RxR are longer and more detailed, describing more landmarks than R2R does. Besides, instructions in RxR no longer describe the shortest path between the starting room to the ending room and the length variance of paths is very large. So, agents cannot simply go directly to the targets and cannot simply use the strong prior of path length to navigate.

\vspace{0pt}
The RxR dataset contains 126,069 navigation instructions, 16,522 trajectories. The average length of each instruction is 78 words.
The instructions of RxR are in three language (\ie English, Hindi, and Telugu). Since our pre-training instructions are in English, here we use English monolingual baseline.

\vspace{0pt}
\subsection{REVERIE}
The task of REVERIE~\cite{reverie} gives a concise, high-level instruction referring a remote object, such as ``Close the kitchen window''. REVERIE requires the agent to follow instructions to navigate and identify the target object in previous unseen environment.

\vspace{3pt}
The REVERIE dataset contains 21,702 instructions. The average length of each instruction is 18 words. The dataset has 4,140 target objects, divided into 489 categories. On average, each target viewpoint has 7 objects with 50 bounding boxes.

\vspace{3pt}
\subsection{Navigation from Dialog History (NDH)}
In the task of NDH~\cite{ndh}, the agent is required to find the target location based on the dialog history, which consists of multiple question-and-answer interactions between the agent and its partners. NDH is much more challenging because the instructions from the dialog history are often ambiguous and unspecified. As a result, agents can hardly navigate to the final location directly.

\vspace{3pt}
The CVDN dataset is used for NDH task, which contains 2050 human-human navigation dialog and over 7K trajectories. 
NDH has three settings: (1) Oracle, which utilizes the shortest path as ground truth observed by the Oracle; (2) Navigator, which uses  the path adopted by human navigator as ground truth; (3) Mixed, which takes the shortest path or the path of human if the human visits the target location.

\begin{table*}[!t]
\renewcommand\arraystretch{1.2}
\small
\centering
\resizebox{\linewidth}{!}{
\begin{tabular}{l |cccc}
\toprule
&\textbf{HOP (proposed)}& \textbf{PREVALENT~\cite{prevalent}}& \textbf{VLN-BERT~\cite{bertvln}}& \textbf{Airbert~\cite{airbert}}\\
\hline
\multirow{3}{*}{\textbf{Dataset}} &Augmented R2R dataset & Augmented R2R dataset & Conceptual Captions~\cite{cc}  & Conceptual Captions~\cite{cc}\\

 &Processed BnB dataset & & Wikipedia and BookCorpus& BnB dataset \\
&                        & &R2R dataset& \\
\hline
\textbf{Visual Input}&Trajectory & Panoramic view (single step) & Trajectory & Trajectory\\
\hline
\multirow{5}{*}{\textbf{Objectives}}& Masked Language Modeling & Masked Language Modeling & Masked Language Modeling& Masked Language Modeling\\
& Action Prediction with History & Action Prediction & Image-Caption matching& Image-Caption matching\\
& Trajectory-Instruction Matching & &Trajectory-Instruction matching & Trajectory-Instruction matching \\
& Trajectory Order Modeling & &&(shuffling loss)\\
& Group Order Modeling\\
\hline
\textbf{Downstream task}& R2R, REVERIE, NDH, RxR & R2R, NDH, HANNA& R2R& R2R, REVEIRIE  \\
\bottomrule
\end{tabular}}
\vspace{-0mm}
\caption{Comparison with related works.}
\label{tab:model_compare}
\vspace{-0mm}
\end{table*}

\subsection{Evaluation Metrics}
\noindent \textbf{SPL} Success weighted by Path Length trades-off SR (Success Rate) against TL (Trajectory Length).\\
\noindent \textbf{nDTW} Normalized dynamic
timewarping penalizes deviations from the reference path.\\
\noindent \textbf{sDTW} Success weighted by normalized Dynamic TimeWarping, constrains nDTW to only successful episodes and effectively captures both success and fidelity.\\
\noindent \textbf{CLS} Coverage weighted by Length measures the overall correspondence between predicted and ground truth trajectories.

\section{Comparison with Related Work}
As shown in table~\ref{tab:model_compare}, we summarize the differences between our method HOP and related VLN pretraining methods, such as PREVALENT~\cite{prevalent}, Airbert~\cite{airbert} and VLN-BERT~\cite{bertvln}.

For visual inputs, we use trajectory instead of a static panoramic image of a single step. In addition to the common Mask Language Modeling (MLM) task and Trajectory-Instruction Matching (TIM) task, we propose two tasks to model temporal order information: Trajectory Ordering Modeling (TOM) and Group Ordering Modeling (GOM). Navigation information is enhanced by introducing the Action Prediction with History (APH) task. Finally, we conduct experiments on four downstream tasks to verify agents from different perspectives.

\section{Results}
As shown in Table~\ref{tab:extra_result_r2r}, we add nDTW and CLS metrics for R2R results.  
\vspace{0pt}

\begin{table}[!h]
\centering
\resizebox{1\linewidth}{!}{
\begin{tabular}{l cc|cc}
\toprule
\multicolumn{1}{l}{\multirow{2}{*}{Methods}}& \multicolumn{2}{c}{R2R Validation Seen} & \multicolumn{2}{c}{R2R Validation Unseen}\\ 
~  &nDTW $\uparrow$& CLS $\uparrow$ &nDTW $\uparrow$& CLS $\uparrow$ \\ 
\midrule
Self-Monitoring~\cite{selfmonitor} & 65.4 & 64.1& 43.7 & 41.5\\
EnvDrop~\cite{envdrop} & 67.2 & 67.2 & 56.7 & 57.0 \\
Syntax~\cite{syntax}& 70.0 & 70.0 & 59.0& 59.0\\
\midrule
HOP & \textbf{76.1} & \textbf{73.3} & \textbf{65.8} &  \textbf{64.5}\\
\bottomrule
\end{tabular}}
\vspace{-1mm}
\caption{
Performance of HOP on R2R.
%
%
}
\vspace{0mm}
\label{tab:extra_result_r2r}
\end{table}

\section{Limitations and Future work}
Apart from the issue of computational resources, one other limitation of our work may be that we did not utilize masked image modeling tasks  for VLN pre-training, such as Masked Region Classification (MRC) task, Masked Region Feature Regression (MRFR) task, etc. These proxy tasks might further improve the performance and generalization of our proposed pre-training methods, especially on downstream VLN tasks that require an agent to locate the target object such as REVERIE. In the future work, we will consider designing more effective masked image modeling tasks for VLN pre-training.

\end{document}